\documentclass[preprint,12pt]{elsarticle}
\usepackage{amssymb}
\usepackage{amsthm}
\usepackage{lineno}
\usepackage{bm}
\usepackage[colorlinks=true, pdfauthor=author]{hyperref}
\usepackage{wrapfig}
\usepackage{booktabs} 
\usepackage{algorithm}
\usepackage{algorithmic}
\usepackage{graphicx}
\usepackage{amsmath}
\usepackage{amssymb}
\usepackage{amsfonts}
\usepackage{multirow}
\usepackage{graphicx}
\usepackage{mathrsfs}
\usepackage{colortbl}
\usepackage{subfigure}
\usepackage{indentfirst}
\usepackage{ulem}
\usepackage{url}
\usepackage{verbatim}
\usepackage{rotating}
\usepackage{adjustbox}
\usepackage{wrapfig}
\usepackage{setspace}

\newtheorem{remark}{Remark}
\graphicspath{{./images/}}

\journal{arXiv}
\begin{document}

\begin{frontmatter}



\title{TabLoRA: Parameter-Efficient Low-Rank Ensemble Learning for Large-Scale Tabular Data}

\author[first]{Jiaqi Luo}
% \cortext[cor1]{Corresponding author}
\ead{jqluo@suda.edu.cn}
\affiliation[first]{organization={School of Mathematical Sciences, Soochow University},
            addressline={No.1 Shizi Street}, 
            city={Suzhou},
            postcode={215006}, 
            state={Jiangsu Province},
            country={China}}

\author[second]{Shixin Xu \corref{cor1}}
\cortext[cor1]{Corresponding author}
\ead{sx59@duke.edu}
\affiliation[second]{organization={Digital innovation research center, Duke Kunshan University},%Department and Organization
            addressline={No.8 Duke Avenue}, 
            city={Kunshan},
            postcode={215000}, 
            state={Jiangsu Province},
            country={China}}

\begin{abstract}
Tabular learning is still dominated by gradient-boosted decision trees (GBDTs), while recent deep learning approaches have become increasingly competitive.
However, applying deep tabular models to large-scale datasets remains challenging, as large sample sizes, high feature dimensionality, or many target classes can introduce substantial computational cost. 
We propose \textbf{TabLoRA}, a parameter-efficient trainable neural ensemble for large-scale tabular learning. 
Instead of using fully independent ensemble backbones, TabLoRA shares a common backbone across predictors and introduces predictor-specific low-rank adaptations, enabling ensemble-style prediction without full parameter duplication. 
Across benchmarks, TabLoRA achieves a favorable balance between predictive performance and practical efficiency compared with GBDT methods and recent deep learning baselines under the same resource constraints. 
Memory analysis and ablation studies further show that the proposed design improves the feasibility of neural ensemble learning while preserving much of the benefit of full ensembles.

% Tabular data remains a fundamental data modality in real-world applications, where gradient-boosted decision trees (GBDT) continue to dominate due to their strong empirical performance. While recent advances have demonstrated the power of deep learning for tabular data, existing approaches either struggle to scale to large datasets or incur substantial computational cost. 
% In this paper, we propose \textbf{TabLoRA}, a parameter-efficient ensemble framework for large-scale tabular learning. Instead of training multiple independent models, TabLoRA constructs an ensemble through a shared backbone network combined with predictor-specific low-rank adaptations. This design allows each predictor to capture distinct representations while sharing the majority of parameters, significantly reducing parameters. 
% Extensive experiments on both classification and regression benchmarks demonstrate that TabLoRA achieves a favorable trade-off between efficiency and performance, outperforming strong baselines including GBDT and recent MLP-based models under comparable resource constraints. 

\end{abstract}


% \begin{highlights}
% \end{highlights}


\end{frontmatter}


\section{Introduction}
\label{s:intro}
 
Tabular data remains a fundamental modality in real-world machine learning applications \cite{jiang2026representation}. 
Historically, gradient-boosted decision trees (GBDTs) \cite{chen2016xgboost,ke2017lightgbm,prokhorenkova2018catboost} have long dominated this domain due to their strong empirical performance and robustness \cite{borisov2022deep,shwartz2022tabular,grinsztajn2022tree}. 
Recently, however, tabular deep learning has become increasingly competitive. 
In particular, in-context learning methods such as TabPFN \cite{hollmann2022tabpfn,hollmann2025accurate,grinsztajn2026tabpfn} demonstrate that deep learning models can match or even outperform GBDT on small- and medium-scale datasets when properly formulated.

Despite this progress, large-scale tabular learning remains computationally challenging. 
Here, large-scale datasets refer to settings where scale may arise from a large number of samples, high feature dimensionality, or many target classes. 
In these regimes, GBDTs become  bottlenecked by tree construction, split search, and class-wise modeling costs, while deep learning models suffer from prohibitive memory and computational footprints during training, making these methods costly or even infeasible.

% Among deep learning approaches, Multi-Layer Perceptrons (MLPs) offer a promising path forward for large-scale learning due to their structural simplicity and high hardware efficiency. 
% While plain MLPs have historically lagged behind GBDT methods, recent studies show that MLP-like models can become highly competitive when combined with the traditional machine learning-induced strategies, like. 
% Among these strategies, ensemble-style prediction is especially effective for improving robustness and predictive performance. 
% However, full neural ensembles require multiple independent predictors and duplicate backbone parameters, leading to parameter growth and increased computational burden as the number of predictors increases.

Among deep learning approaches, MLP-based models provide a simple and computationally efficient foundation, making them naturally amenable to large-scale settings. Recent work further shows that incorporating classical machine learning structures \cite{luo2024ncart,gorishniy2024tabm,ye2024modern} can significantly enhance their performance. Compared to more complex architectures, such designs offer practical efficiency and are easier to scale to large datasets.

However, even within this design space, a fundamental challenge remains. Highly expressive MLP-based models often incur substantial memory overhead that grows with feature dimensionality and dataset size, while memory-efficient designs tend to suffer from limited representation capacity. This reveals an inherent trade-off between scalability and expressivity in large-scale tabular learning, which remains insufficiently addressed by existing approaches.

To overcome this limitation, we propose \textbf{TabLoRA}, a parameter-efficient ensemble framework that combines shared backbone learning with predictor-specific low-rank adaptations. 
Each predictor is modeled as a low-rank perturbation of shared weights, allowing the model to approximate the behavior of deep ensembles without incurring linear growth in memory cost. 
In addition, we introduce lightweight feature transformations that generate multiple input representations, further enhancing diversity across predictors without increasing feature dimensionality. 
Experimental results on large-scale dataset benchmarks show that TabLoRA achieves strong predictive performance and a favorable practical performance–efficiency trade-off compared with the state-of-the-art baselines. Further ablation studies demonstrate that TabLoRA significantly reduces the parameters of neural ensemble learning  while preserving the predictive capability of full ensembles.

The main contributions of this paper are summarized as follows:
\begin{itemize}
    \item We propose TabLoRA, a parameter-efficient trainable neural ensemble for large-scale tabular learning.
    \item We design a shared-backbone, low-rank adaptation mechanism that enables predictor-specific specialization without full ensemble parameter duplication.
    \item We empirically show that TabLoRA achieves a favorable performance--efficiency trade-off on large-scale tabular benchmarks, with ablations validating the role of the feature adapter and low-rank ensemble parameterization.
\end{itemize}

\section{Related Work}
\label{s:rel}

\subsection{Deep Learning for Tabular Data}

This section gives an overview of relevant concepts from prior research on deep learning for tabular data. We categorize deep learning models into three types based on their network structure: tree-induced networks, transformer-based networks, and other specialized models.

\paragraph{MLP-based Architectures}
MLP-based models form one of the most practical directions in tabular deep learning. 
Although plain MLPs often underperform GBDT methods, recent studies show that their performance can be substantially improved by incorporating techniques from classical machine learning, better regularization, feature encoding, retrieval, and ensemble learning. 
Early efforts improve the training and generalization of MLPs through regularization and carefully designed default settings~\cite{shavitt2018regularization,kadra2021well,holzmuller2024better}. 
Another line of work focuses on improving input representations. 
For example, MLP-PLR~\cite{gorishniy2022embeddings} introduces numerical feature encoding methods to better represent continuous variables, while TabR~\cite{gorishniy2024tabr} augments MLP-based prediction with retrieval mechanisms to improve robustness and predictive performance. 
Other methods integrate classical learning principles more directly into neural architectures. 
ModernNCA~\cite{ye2024modern} incorporates Neighbourhood Component Analysis into an MLP-based framework, whereas TabM~\cite{gorishniy2024tabm} uses MLPs as base learners and adopts BatchEnsemble and numerical feature encoding to construct multiple diverse predictors efficiently. 
Similarly, NCART~\cite{luo2024ncart} combines MLPs with decision-tree-based ensemble learning. 
Together, these studies suggest that MLP-like backbones remain competitive when equipped with suitable tabular-specific mechanisms.

\paragraph{Transformer-based Architectures}
Transformer-based models provide another important direction for tabular learning by explicitly modeling interactions among features or samples through attention mechanisms. 
Inspired by the success of Transformers~\cite{vaswani2017attention}, early methods adapt self-attention to heterogeneous tabular inputs. 
TabTransformer~\cite{huang2020tabtransformer} maps categorical features into contextual embeddings using self-attention, improving robustness to missing or noisy values and offering a degree of interpretability. 
FT-Transformer~\cite{gorishniy2021revisiting} provides a more direct adaptation by tokenizing both numerical and categorical features and feeding them jointly into Transformer blocks. 
Beyond feature-wise attention, several methods introduce more structured attention mechanisms. 
TabNet~\cite{arik2021tabnet} adopts a sequential decision process with soft instance-wise feature selection, while SAINT~\cite{somepalli2021saint} combines feature-wise self-attention with inter-sample attention and further uses self-supervised contrastive pre-training. 
NPT~\cite{kossen2021self} treats the entire dataset as input and uses attention between data points to model sample-level relationships. 
More recent models further refine Transformer-based tabular learning with specialized interaction modules or structural information. 
ExcelFormer~\cite{chen2024can} alternates between attention modules for feature interaction and embedding updates, and T2GFormer~\cite{yan2023t2g} explores Transformer-based modeling with graph-based structures. 
These methods demonstrate the flexibility of attention mechanisms for tabular data, although their computational cost can become a concern in large-scale settings.

\paragraph{Other Specific Architectures}
Beyond MLP- and Transformer-based models, many studies design specialized architectures to address the structural heterogeneity of tabular data. 
Unlike images or language, tabular data does not have a universal spatial or sequential structure, which makes it difficult to directly transfer standard deep learning architectures. 
One strategy is to transform tabular data into other modalities, such as images or text, and then apply models designed for those domains~\cite{sun2019supertml,yin2020tabert,hegselmann2023tabllm}. 
Another strategy is to design task-specific neural architectures that better capture feature interactions and high-level representations. 
For instance, DANets~\cite{chen2022danets} and TabCaps~\cite{chen2022tabcaps} introduce specialized structures for representation learning, while NODE~\cite{popov2019neural} integrates differentiable oblivious decision trees into an ensemble-style neural model. 
In addition, self-supervised learning methods~\cite{yoon2020vime,ucar2021subtab,hajiramezanali2022stab} and transfer learning approaches~\cite{wang2022transtab,levin2022transfer} have been introduced to improve representation learning and downstream predictive performance. 
These works reflect the diversity of architectural designs for tabular deep learning.

\paragraph{Pre-trained Models}
Pre-trained tabular models have recently emerged as a promising direction, aiming to transfer knowledge across datasets and adapt to new tabular tasks with limited task-specific training. 
A representative example is TabPFN~\cite{hollmann2022tabpfn,hollmann2025accurate}, which demonstrates the effectiveness of in-context learning for small-scale tabular prediction and has recently been extended toward more scalable settings~\cite{grinsztajn2025tabpfn,grinsztajn2026tabpfn}. 
Following this direction, several works further improve scalability, generality, or task coverage. 
TabDPT~\cite{ma2026tabdpt} combines retrieval techniques with self-supervised learning to train tabular foundation models, while TabICL~\cite{qu2025tabicl,qu2026tabiclv2} employs a column-then-row attention mechanism to address the scalability limitations of TabPFN. 
LimiX~\cite{zhang2025limix,wang2026limix} further broadens the scope by using a single model for multiple tabular tasks, including classification, regression, missing-value imputation, feature selection, and sample selection. 
These methods show the potential of pre-trained tabular predictors, although their inference and adaptation costs remain important considerations in large-scale applications.
% TabSWIFT is a lightweight tabular foundation model that remains competitive with stronger pre-trained tabular models while improving inference efficiency.

\subsection{Parameter-Efficient Fine-Tuning}

Parameter-efficient fine-tuning (PEFT) \cite{ding2023parameter} aims to adapt pretrained models to downstream tasks by updating only a small number of task-specific parameters. Instead of fine-tuning the entire backbone, PEFT methods usually freeze most pretrained weights and introduce lightweight trainable components, such as adapters and low-rank updates. 

Adapter tuning is a representative PEFT strategy. Houlsby et al.~\cite{houlsby2019parameter} introduced bottleneck adapters, where small trainable modules are inserted into pretrained networks while the backbone remains fixed.
Another important PEFT method is low-rank adaptation. LoRA~\cite{hu2022lora} freezes pretrained weights and injects trainable low-rank matrices into selected layers, representing weight updates through low-rank decompositions. This greatly reduces the number of trainable parameters while maintaining competitive performance. Several extensions further improve LoRA's flexibility and efficiency. AdaLoRA~\cite{zhang2023adalora} dynamically allocates the parameter budget across weight matrices, whereas QLoRA~\cite{dettmers2023qlora} combines LoRA with three innovations to enable memory-efficient fine-tuning of large language models.

\section{Methodology}
\label{s:method}

\subsection{Overview}
\begin{figure}[!ht]
    \centering
    \includegraphics[width=\linewidth]{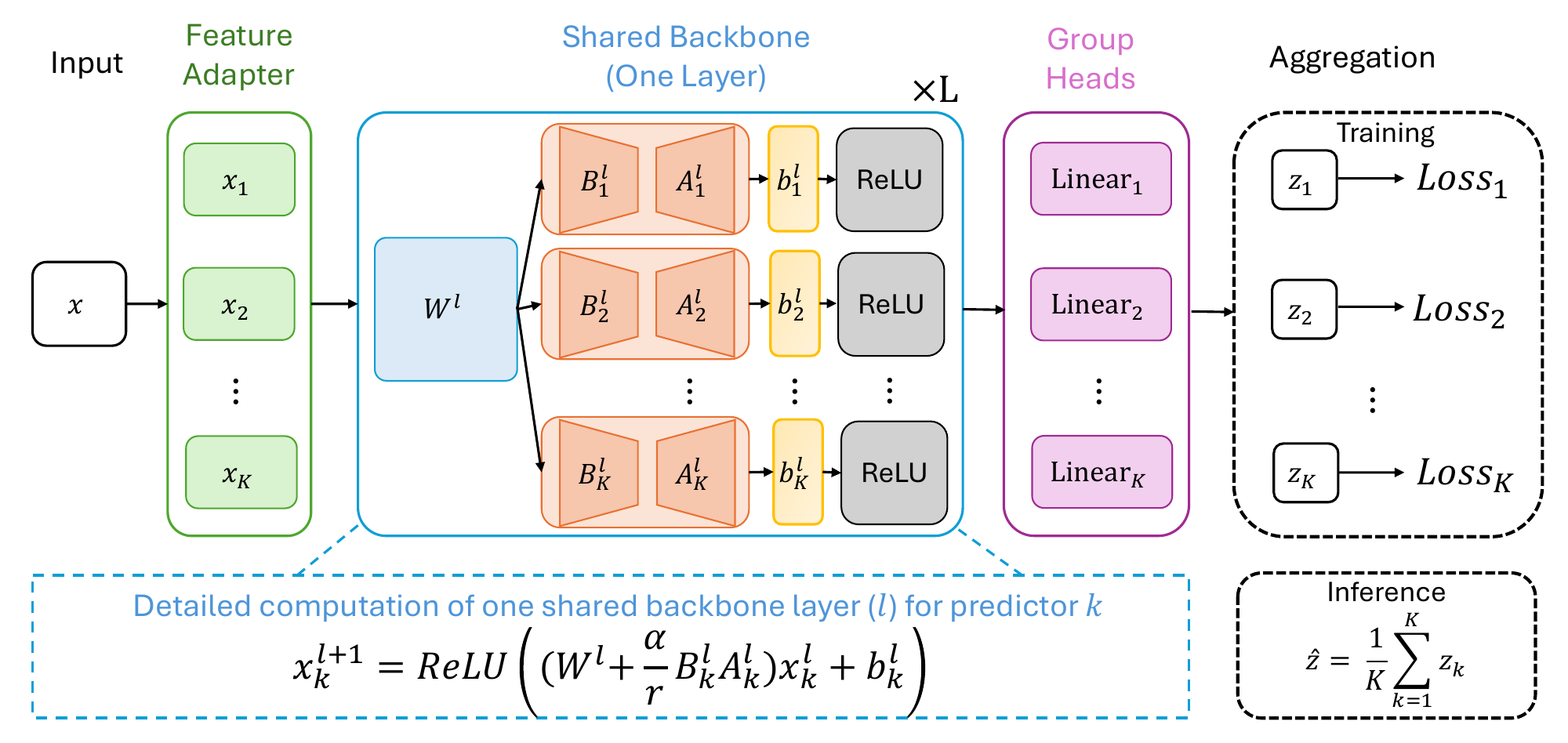}
    \caption{Overview of TabLoRA. 
    The model combines predictor-specific feature adaptation, shared backbone learning with low-rank adaptations, and ensemble aggregation.}
    \label{f.framework}
\end{figure}

We propose a parameter-efficient ensemble framework for tabular learning.
As illustrated in Fig.~\ref{f.framework}, our method constructs multiple predictors through a shared backbone with low-rank adaptations, enabling efficient ensemble modeling.

Given an input sample $x \in \mathbb{R}^{D}$, the model produces $K$ predictions $\{f_k(x)\}_{k=1}^{K}$ in parallel,
which are later aggregated during training and inference.
The overall architecture consists of three components:
(1) a feature adapter that generates $K$ representations,
(2) a shared backbone with low-rank adaptations,
and (3) a group-wise prediction head.

\subsection{Feature Expansion via Group Adapter}

To construct multiple predictors efficiently, we first transform the input feature vector into $K$ parallel representations.
Given an input feature $x \in \mathbb{R}^{D}$, we define
\begin{equation}
   x_k = x + \phi_k(x), 
\end{equation}
where $\phi_k(\cdot)$ is an adapter~\cite{houlsby2019parameter}:
\begin{equation}
   \phi_k(x) = W^{(k)}_{\text{up}} \, \sigma \left( W^{(k)}_{\text{down}} x \right). 
\end{equation}
Here, $W^{(k)}_{\text{up}} \in \mathbb{R}^{D \times r}$ and  $W^{(k)}_{\text{down}} \in \mathbb{R}^{r \times D}$ are trainable, $\sigma(\cdot)$ is a non-linear activation.
This results in $K$ feature representations
\begin{equation}
    X = [x_1, x_2, \dots, x_K] \in \mathbb{R}^{K \times D}.
\end{equation}
This design introduces diversity across predictors while maintaining a small parameter footprint due to the low-rank bottleneck.

\subsection{Parameter-Efficient Ensemble via Low-Rank Adaptation}
Instead of training \(K\) independent networks, we use a shared backbone with per-predictor low-rank adaptations \cite{hu2022lora}. 
Unlike conventional LoRA, which freezes a pretrained backbone for parameter-efficient fine-tuning, TabLoRA uses low-rank adaptation as an ensemble parameterization. 
Specifically, the shared backbone and the predictor-specific low-rank adaptations are trained jointly from scratch for tabular ensemble learning.
For a linear layer with weight $W \in \mathbb{R}^{d_{\text{out}} \times d_{\text{in}}}$, the output for predictor $k$ is defined as
\begin{equation}
\label{e.lora}
    h_k = Wx_k + \frac{\alpha}{r} \cdot B_kA_kx_k , k=1,2,\cdots, K,
\end{equation}
where $A_k \in \mathbb{R}^{r \times d_{\text{in}}}$, $B_k \in \mathbb{R}^{d_{\text{out}} \times r}$, $r \ll d_{\text{in}}, d_{\text{out}}$ is the low-rank dimension, and $\alpha$ is a scaling factor.

The first term corresponds to the shared backbone, while the second term introduces predictor-specific low-rank perturbations.
This formulation enables the model to approximate multiple predictors with significantly fewer parameters than independent ensembles.

The network is implemented as a multilayer MLP. All linear layers in the network are replaced with the low-rank ensemble backbones described above.
Given the $K$ feature representations $x_k^l$ of the layer $l$, the backbone processes them in parallel
\begin{equation}
\label{e.backbone}
    x_{k}^{l+1} = ReLU \left( W^l x_k^l + \frac{\alpha}{r}\cdot B_k^lA_k^lx_k^l+b^l \right), k=1,2,\cdots, K, 
\end{equation}
where $b_l$ is the shared bias and $ReLU$ is the relu activative function. 

\begin{remark}
If $\alpha=0$ and $K=1$, then the network degenerates to the vanilla MLP.
\end{remark}
% \begin{equation}
%     H' = \text{Backbone}(H) \in \mathbb{R}^{K \times H}.
% \end{equation}

\subsection{Group Prediction Head}

To generate predictions for all ensemble members, we use a group-wise linear head
\begin{equation}
   z_k = W_k^{head} x_k^L + b_k^{head}, 
\end{equation}
where each predictor has its own head parameters $W_k^{head}$ and $b_k^{head}$. The final output is
\begin{equation}
    Z = [z_1, z_2, \dots, z_K].
\end{equation}

\subsection{Training and Inference}
During training, all predictors are treated equally by optimizing the average loss over the $K$ outputs:
\begin{equation}
    \mathcal{L}_{\mathrm{total}}
    =
    \frac{1}{K}
    \sum_{k=1}^{K}
    \mathcal{L}(z_k,y),
\end{equation}
where $y$ is the ground-truth label. For classification, $\mathcal{L}$ is the cross-entropy loss applied directly to the logits $z_k$. For regression, $\mathcal{L}$ is a regression loss such as mean squared error. 

During inference, the $K$ predictor outputs are aggregated by averaging. For regression, the final prediction is
\begin{equation}
    \hat{y}
    =
    \frac{1}{K}
    \sum_{k=1}^{K} z_k .
\end{equation}
For classification, we aggregate the logits and then apply the softmax function:
\begin{equation}
    \hat{p}
    =
    \mathrm{softmax}
    \left(
    \frac{1}{K}
    \sum_{k=1}^{K} z_k
    \right),
\end{equation}
where $\hat{p}$ is the predicted class probability vector. 
% The predicted class is then given by
% \begin{equation}
%     \hat{y}
%     =
%     \arg\max_{c} \hat{p}_c .
% \end{equation}

\subsection{Complexity Analysis}
\label{s.complex}

We compare the parameter complexity of TabLoRA with a standard deep ensemble. 
For simplicity, we focus on the backbone parameters and omit lower-order terms such as biases and prediction heads. 
Consider an $L$-layer MLP backbone with input dimension $d$ and hidden dimension $h$. 
A standard deep ensemble with $K$ independent predictors duplicates the entire backbone for each predictor, resulting in
\begin{equation}
    P_{\mathrm{full}}
    =
    K\big(dh + (L-1)h^2\big).
\end{equation}

In contrast, TabLoRA shares the main backbone across all predictors. 
The shared backbone contains
\begin{equation}
    P_{\mathrm{shared}}
    =
    dh + (L-1)h^2
\end{equation}
parameters. 
For each predictor, TabLoRA introduces low-rank adaptations to the backbone weights. 
For the input layer, whose weight matrix has size $h \times d$, the low-rank adaptation introduces $r(d+h)$ parameters. 
For each hidden layer, whose weight matrix has size $h \times h$, the low-rank adaptation introduces $2hr$ parameters. 
Therefore, the predictor-specific low-rank parameters for all $K$ predictors are
\begin{equation}
    P_{\mathrm{LoRA}}
    =
    K\Big(r(d+h) + 2(L-1)hr\Big)
    =
    Kr\big(d + (2L-1)h\big),
\end{equation}
where $r \ll h$ is the adaptation rank.

The total backbone-related parameter complexity of TabLoRA is therefore
\begin{equation}
    P_{\mathrm{TabLoRA}}
    =
    dh + (L-1)h^2
    +
    Kr\big(d + (2L-1)h\big).
\end{equation}

Compared with the standard ensemble, TabLoRA avoids duplicating the full backbone for each predictor. 
The ensemble-specific parameter growth is reduced from $Kh\big(d + (L-1)h\big)$ to $Kr\big(d + (2L-1)h\big)$.
Since $r \ll h$, the predictor-specific cost in TabLoRA is much smaller than maintaining $K$ independent backbones. 
Thus, TabLoRA provides a parameter-efficient ensemble parameterization while preserving multiple predictor-specific adaptations.

% \subsection{Complexity Analysis}
% \label{s.complex}
% We compare the parameter complexity of the proposed method with standard deep ensembles.
% A conventional ensemble with $K$ independent models requires:
% \begin{equation}
%    O\big(K \cdot (dh + (L-1)h^2)\big) \approx O(K h^2), 
% \end{equation}
% where $d$ is the input dimension, $h$ is the hidden dimension ($d_{\text{out}}=d_{\text{in}}$ in \eqref{e.lora}), and $L$ is the number of layers.

% In contrast, the proposed method shares a common backbone and introduces low-rank adaptations.
% The backbone requires
% \begin{equation}
%   O(dh + (L-1)h^2),  
% \end{equation}
% while the low-rank components introduce an additional
% \begin{equation}
%   O(2\cdot K \cdot L \cdot h \cdot r),  
% \end{equation}
% where $r \ll h$ is the rank of the adaptation.
% The total parameter complexity becomes
% \begin{equation}
%     O(dh + Lh^2 + K L h r).
% \end{equation}

% Compared to standard ensembles, the proposed formulation reduces the parameter growth from $O(K h^2)$ to $O(h^2 + K h r)$.
% Since $r \ll h$, this leads to a substantial reduction in parameter complexity while maintaining multiple predictors.

\section{Experimental Setup}
\label{s:exps}

\subsection{Datasets and Baselines}
We conduct experiments on 16 tabular datasets from OpenML\footnote{\url{https://www.openml.org/}}, covering both classification and regression tasks. The classification datasets include 5 binary and 7 multi-class problems. Dataset details can be found in \ref{a.data}.
We compare the proposed method with three widely used GBDT models and five representative deep tabular learning baselines. The GBDT baselines include XGBoost \cite{chen2016xgboost}, CatBoost \cite{prokhorenkova2018catboost}, and LightGBM \cite{ke2017lightgbm}. The neural baselines include MLP, RealMLP \cite{holzmuller2024better}, NCART \cite{luo2024ncart}, TabM \cite{gorishniy2024tabm}, and TabPFN \cite{grinsztajn2026tabpfn}. 

\subsection{Evaluation Metrics}
We evaluate model performance using \textbf{AUC} (Area Under the Curve) for binary classification and \textbf{Acc.} (Accuracy) for multi-class classification. For regression tasks, we use \textbf{MSE} (Mean Squared Error). Since these metrics have different directions and scales, 
we define a unified relative improvement metric based on gap reduction with respect to MLP.
For each dataset $d$ and method $m$, we define the relative improvement as
\begin{equation}
\label{e.improv}
I_{m,d} =
\begin{cases}
\dfrac{(1-S_{\mathrm{MLP},d}) - (1-S_{m,d})}
      {1 - S_{\mathrm{MLP},d}},
& \text{classification}, \\[1.2em]
\dfrac{\mathrm{MSE}_{\mathrm{MLP},d} - \mathrm{MSE}_{m,d}}
      {\mathrm{MSE}_{\mathrm{MLP},d}},
& \text{regression},
\end{cases}
\end{equation}
where $S_{m,d}$ denotes AUC for binary classification and Acc. for multi-class classification.
% For classification, this metric measures the relative reduction of the gap to the best possible score; for regression, it corresponds to the relative MSE reduction over MLP. 
A positive value of $I_{m,d}$ indicates that method $m$ improves over MLP, while a negative value indicates worse performance.

\subsection{Implementation Details}
To ensure fair and reproducible evaluation, we adopt a standardized experimental protocol across all datasets and methods. Each dataset is randomly split into training 80\% and testing 20\% sets using stratified sampling to preserve class distribution. Within the training set, 10\% is further reserved as a validation set for hyperparameter tuning.
Hyperparameters are optimized using Optuna \cite{akiba2019optuna} with a budget of 10 trials per method. After tuning, models are retrained on the full training set and evaluated on the test set.
To account for randomness, we repeat the entire process five times with different random seeds and report the mean and standard deviation of all metrics. 

For neural network models, we train for up to 200 epochs using the Adam optimizer with a learning rate of 0.001. For GBDT models, the number of boosting iterations is set to 200. Early stopping is applied with a patience of 20 epochs or iterations. The training batch size is set to 1024, and the validation batch size is set to 256.
For each algorithm and each training process, we run the algorithm for up to 10 hours to prevent excessively long runtimes.
All experiments are conducted on a workstation equipped with an Intel Core i9-14900KF CPU, 128GB RAM, and a  24G NVIDIA-4090 GPU.
% More details about the datasets and the hyperparameters can be found in \ref{a.data} and \ref{a.params}.
More details about the hyperparameters can be found in \ref{a.params}.

\section{Results}
\subsection{Main Results on Large-Scale Datasets}

Table~\ref{T.mainresults} summarizes the predictive performance on 16 large-scale datasets. 
The results also reveal clear differences in robustness and feasibility across methods. 
TabPFN achieves the largest number of best results when it runs successfully, especially on several multi-class and regression datasets. 
However, it suffers from multiple OOM failures on large-scale datasets, reflecting the computational burden of applying in-context tabular prediction under large-scale data regimes. 
TabM also obtains strong results on some datasets, but fails on several large-scale datasets and gives the worst performance on multiple tasks. This is related to its feature embedding mechanism, which can increase the cost of intermediate representations. 
NCART also faces OOM failures on some datasets, since it constructs ensemble-style neural tree models and can become expensive when the feature dimension is high. 
RealMLP improves over the standard MLP with better training strategies and default configurations. 
As a single-network model, it avoids OOM failures but its predictive performance is less competitive than stronger ensemble-style or foundation-model baselines. 

GBDT models remain reliable baselines, but their overall ranks are worse than TabLoRA. 
XGBoost and CatBoost also encounter OOM failures on some large-scale datasets. 
These failures mainly occur on high-dimensional datasets, suggesting that the memory cost of tree construction is strongly affected by feature-wise split statistics and intermediate buffers, rather than by sample size alone. 
In contrast, LightGBM does not encounter OOM failures in our experiments, which may be attributed to its histogram-based learning strategy and memory-efficient implementation for large-scale tree construction. 

This observation also indicates that different forms of scale affect different model families differently. High feature dimensionality is particularly challenging for tree-based and feature-embedding methods, while in-context tabular prediction can also be sensitive to large sample sizes and context construction.
The standard MLP baseline performs poorly overall, confirming that simply scaling a plain neural network is insufficient for large-scale tabular prediction. 
In comparison, TabLoRA obtains the best result on four datasets and is not the worst method on any dataset, showing a strong balance between predictive performance and practical feasibility.

\begin{figure}[!ht]
    \centering
    \includegraphics[width=0.88\linewidth]{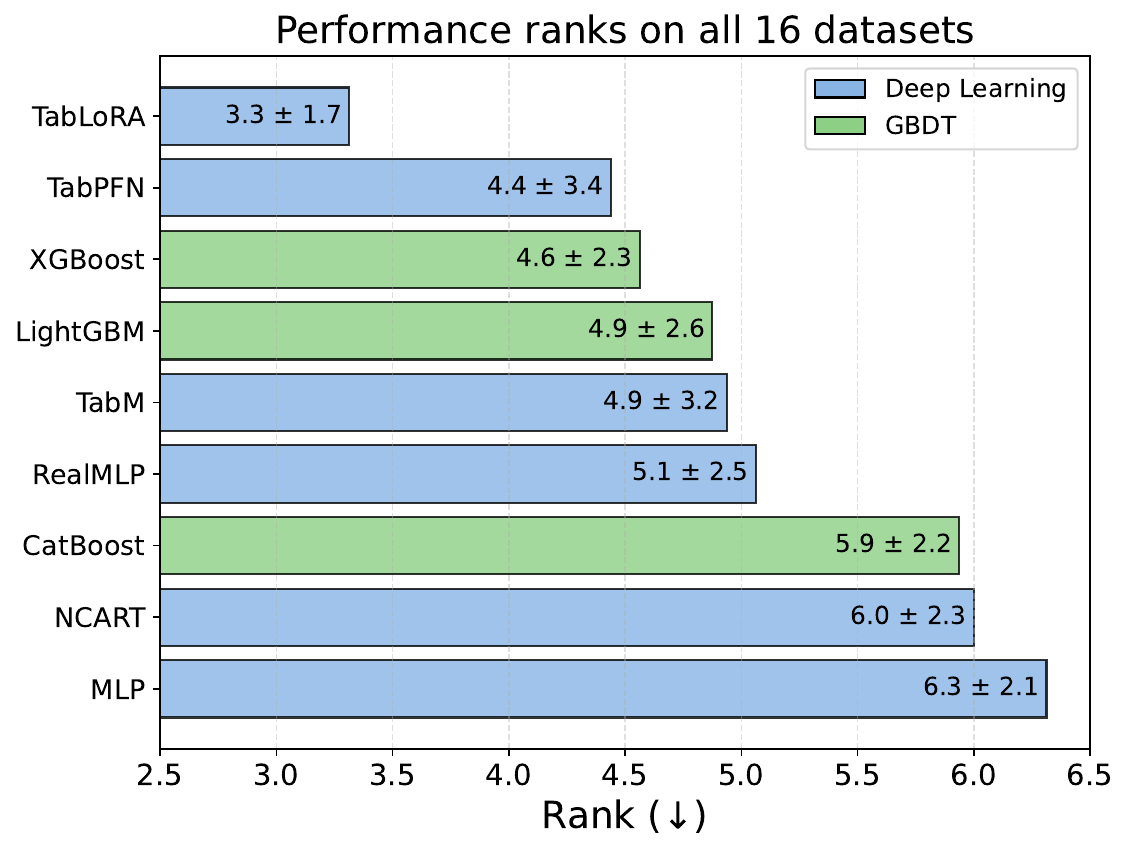}
    \caption{Average performance ranks on large-scale datasets. Lower rank indicates better performance. Blue bars denote deep learning-based methods, and green bars denote GBDT methods.}
    \label{f.rank}
\end{figure}

\begin{figure}[!ht]
    \centering
    \includegraphics[width=0.88\linewidth]{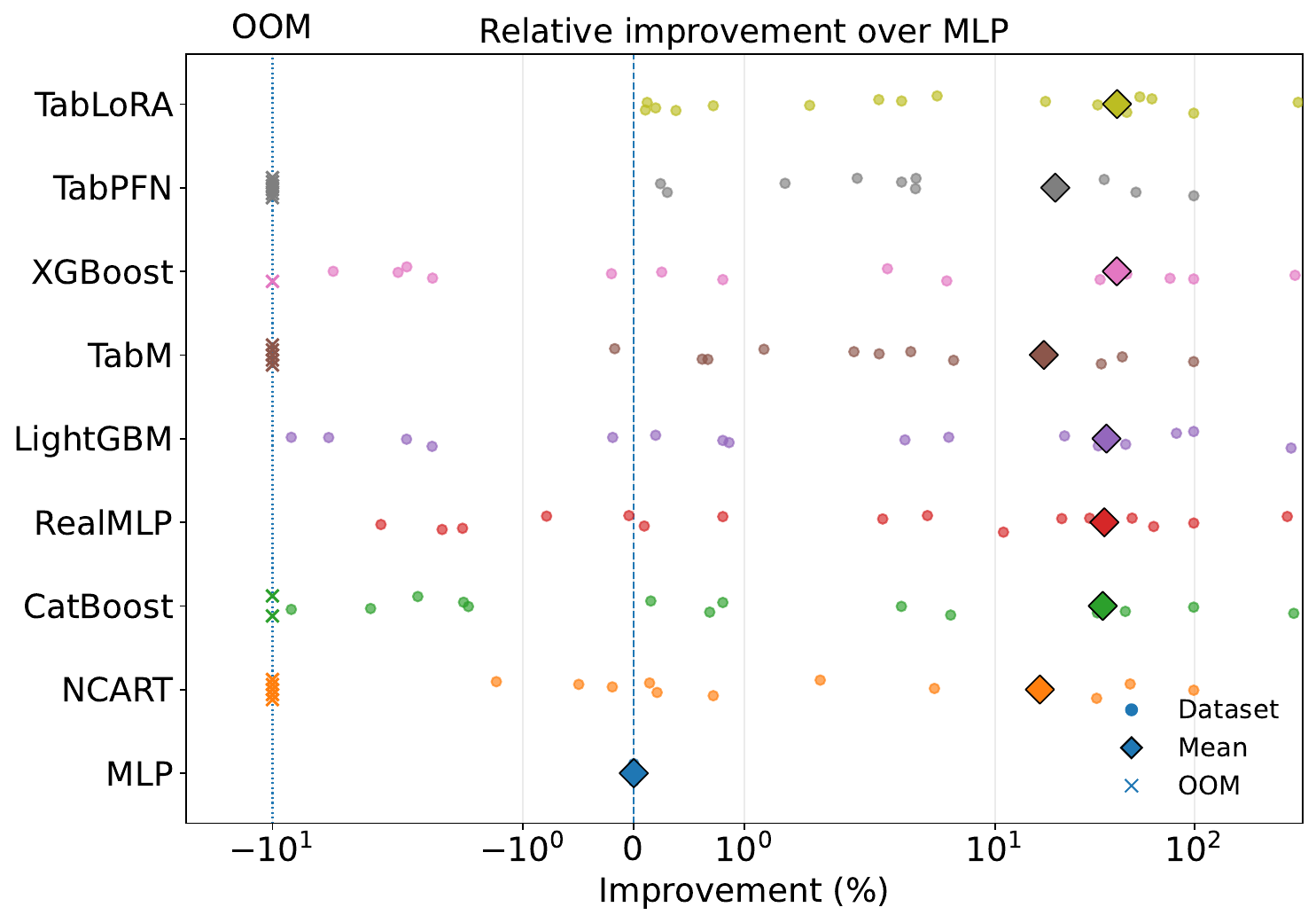}
    \caption{Dataset-level relative improvement over MLP on large-scale datasets. Each point represents one dataset, and the diamond marker denotes the mean improvement over successful runs. Cross markers indicate \textit{OOM} cases, which are shown separately and are not assigned numerical improvement values.}
    \label{f.improv}
\end{figure}

Fig.~\ref{f.rank} summarizes the overall ranking across all large-scale datasets. 
TabLoRA achieves the best average rank among all compared methods, indicating that its advantage is not limited to a few individual datasets but is consistent across different task types. 
Fig.~\ref{f.improv} further provides a dataset-level view of relative improvement over MLP. 
All TabLoRA points lie on the positive side, showing that TabLoRA consistently improves over the plain MLP baseline. 
TabPFN also shows strong positive improvements on successful runs, but its OOM cases limit its practical applicability.
In contrast, several competing methods either show larger variability or suffer from OOM failures on large-scale datasets. 
Together with the main result table, these figures show that TabLoRA achieves strong and stable predictive performance while maintaining good feasibility across the evaluated large-scale datasets.

\begin{sidewaystable}
\renewcommand\arraystretch{1.3}
    \centering
    \caption{
    Mean $\pm$ standard deviation results of nine models on 16 large-scale datasets over five random runs. 
    The best mean result on each dataset is highlighted in \textbf{bold}. 
    For classification tasks, higher AUC or accuracy is better; for regression tasks, lower MSE is better. 
    \textit{OOM} indicates an out-of-memory failure under the same GPU budget.
    }
    \label{T.mainresults}
    \begin{adjustbox}{width=\textwidth}
    \begin{tabular}{lccccccccc}
    \toprule[2pt]
        Dataset & XGBoost & CatBoost & LightGBM & MLP & RealMLP & NCART & TabM & TabPFN & TabLoRA \\
        
    \midrule[1.5pt]
        \multicolumn{10}{c}{Binary Classification (AUC $\uparrow$)}\\
    credit-g & 88.89$\pm$0.18 & 89.11$\pm$0.14 & 89.00$\pm$0.11 & 84.13$\pm$0.66 & 87.94$\pm$0.25 & 88.26$\pm$0.16 & \textbf{89.28$\pm$0.07} & 87.47$\pm$0.11 & 88.39$\pm$0.18 \\
    road-safety & 88.94$\pm$1.07 & 88.26$\pm$0.47 & 88.48$\pm$0.37 & 66.83$\pm$2.31 & 86.43$\pm$1.20 & 88.07$\pm$0.18 & 89.25$\pm$0.11 & \textbf{90.01$\pm$0.14} & 88.34$\pm$0.22 \\
    Epsilon & 94.11$\pm$0.10 & 94.67$\pm$0.10 & 94.40$\pm$0.10 & 96.13$\pm$0.02 & 96.22$\pm$0.06 & \textit{OOM} & \textit{OOM} & \textit{OOM} & \textbf{96.23$\pm$0.03} \\
    vehicleNorm & 92.48$\pm$0.10 & 92.39$\pm$0.12 & 92.43$\pm$0.10 & 92.25$\pm$0.13 & 92.21$\pm$0.31 & 92.38$\pm$0.17 & \textbf{92.86$\pm$0.08} & 92.47$\pm$0.09 & 92.43$\pm$0.07 \\
    Higgs & 82.16$\pm$0.41 & 82.05$\pm$0.23 & 81.90$\pm$0.10 & 83.66$\pm$0.22 & 83.01$\pm$0.84 & 83.50$\pm$0.14 & \textbf{84.17$\pm$0.11} & 83.91$\pm$0.05 & 83.76$\pm$0.13 \\
    
    \midrule[1.5pt]
        \multicolumn{10}{c}{Multiclass Classification (Acc. $\uparrow$)}\\
    covertype & 90.99$\pm$4.93 & 88.07$\pm$2.22 & 88.07$\pm$3.49 & 95.69$\pm$0.47 & 92.98$\pm$1.78 & 95.22$\pm$0.19 & 97.57$\pm$0.07 & \textbf{97.62$\pm$0.06} & 96.05$\pm$0.27 \\
    robert & 50.43$\pm$0.85 & \textit{OOM} & \textbf{52.05$\pm$0.48} & 29.20$\pm$4.24 & 46.77$\pm$1.21 & \textit{OOM} & \textit{OOM} & \textit{OOM} & 44.18$\pm$1.59 \\
    CIFAR-100 & 27.00$\pm$0.31 & \textit{OOM} & 23.55$\pm$1.32 & 19.44$\pm$1.43 & 23.42$\pm$0.65 & \textit{OOM} & \textit{OOM} & \textit{OOM} & \textbf{30.70$\pm$0.56} \\
    Kuzushiji-49 & 89.15$\pm$1.00 & 88.36$\pm$0.60 & 86.55$\pm$2.40 & 91.27$\pm$0.26 & 89.71$\pm$0.72 & 91.46$\pm$0.12 & 93.64$\pm$0.18 & \textbf{94.34$\pm$0.06} & 93.63$\pm$0.36 \\
    isolet & 95.37$\pm$0.84 & 94.15$\pm$0.58 & 95.38$\pm$0.85 & 95.56$\pm$0.44 & 94.10$\pm$1.37 & \textbf{97.15$\pm$0.27} & 96.67$\pm$0.22 & 96.85$\pm$0.45 & 97.06$\pm$0.37 \\
    SVHN & \textit{OOM} & 75.97$\pm$0.44 & 76.10$\pm$0.99 & 75.46$\pm$4.61 & 83.66$\pm$1.87 & \textit{OOM} & \textit{OOM} & \textit{OOM} & \textbf{88.78$\pm$0.34} \\
    eating & 57.88$\pm$3.88 & 57.25$\pm$2.56 & 55.98$\pm$3.02 & 14.60$\pm$0.42 & 54.18$\pm$5.31 & \textit{OOM} & \textit{OOM} & \textit{OOM} & \textbf{59.47$\pm$3.10} \\
    
    \midrule[1.5pt]
    \multicolumn{10}{c}{Regression (MSE $\downarrow$)}\\
    breastTumor & 88.15$\pm$0.61 & 87.69$\pm$0.57 & 87.56$\pm$0.60 & 90.77$\pm$0.69 & 88.29$\pm$0.70 & 91.90$\pm$1.11 & 87.34$\pm$0.69 & \textbf{87.15$\pm$0.61} & 87.68$\pm$0.79 \\
    Yolanda & 79.20$\pm$1.40 & 80.31$\pm$0.71 & 80.08$\pm$0.95 & 145.87$\pm$77.51 & 75.00$\pm$0.93 & 76.57$\pm$0.61 & 82.60$\pm$9.24 & \textbf{71.78$\pm$0.61} & 79.20$\pm$2.39 \\
    SafeDriver \small{(${\times}10^{-3}$)} & \textbf{34.59$\pm$0.45} & 34.59$\pm$0.46 & \textbf{34.59$\pm$0.45} & 34.87$\pm$0.41 & 34.59$\pm$0.46 & 34.62$\pm$0.46 & 34.93$\pm$0.47 & \textit{OOM} & 34.62$\pm$0.45 \\
    year & 78.09$\pm$1.38 & 79.37$\pm$0.85 & 79.42$\pm$0.82 & 6841.00$\pm$13448.61 & 73.21$\pm$0.48 & 75.22$\pm$0.41 & 79.93$\pm$6.24 & \textbf{69.04$\pm$0.49} & 75.49$\pm$0.84 \\

    \midrule[1.5pt]
    Mean rank & 4.6 & 5.9 & 4.9 & 6.3 & 5.1 & 6.0 & 4.9 & 4.4 & \textbf{3.3}\\
    Best/Worst & 1/1 & 0/3 & 2/3 & 0/4 & 0/2 & 1/6 & 3/5 & 6/6 & \textbf{4/0}\\
    \bottomrule[2pt]
    \end{tabular}
    \end{adjustbox}
\end{sidewaystable}

\subsection{Practical Efficiency Analysis}

We further analyze the practical GPU memory usage of deep learning-based methods. 
For each method, we compute the GPU memory multiple relative to MLP and compare it with the average performance rank across the 16 large-scale datasets. 
Fig.~ref{f.bubble} and Fig.~ref{f.gpu} summarize the performance--memory trade-off and the dataset-level memory behavior, respectively. 
The full numerical results are provided in Table~\ref{T.memoryresults} in Appendix~\ref{a.results}.

Fig.~\ref{f.bubble} shows that TabLoRA achieves a favorable balance between predictive performance and average GPU memory usage. 
Compared with other neural baselines, TabLoRA obtains the lowest average rank while keeping the mean GPU memory multiple relatively low. 
In contrast, TabM and TabPFN require substantially higher average GPU memory and suffer from several OOM failures, although they can be competitive on datasets where they successfully run. 
RealMLP is a single-network model, but its average memory usage is still higher than that of TabLoRA, which may be related to its numerical feature embedding mechanism. 
NCART has an average memory usage close to TabLoRA on successful runs, but it also encounters multiple OOM failures, indicating that a low average memory multiple does not necessarily imply stable feasibility across datasets.

\begin{figure}[!ht]
    \centering
    \includegraphics[width=0.95\linewidth]{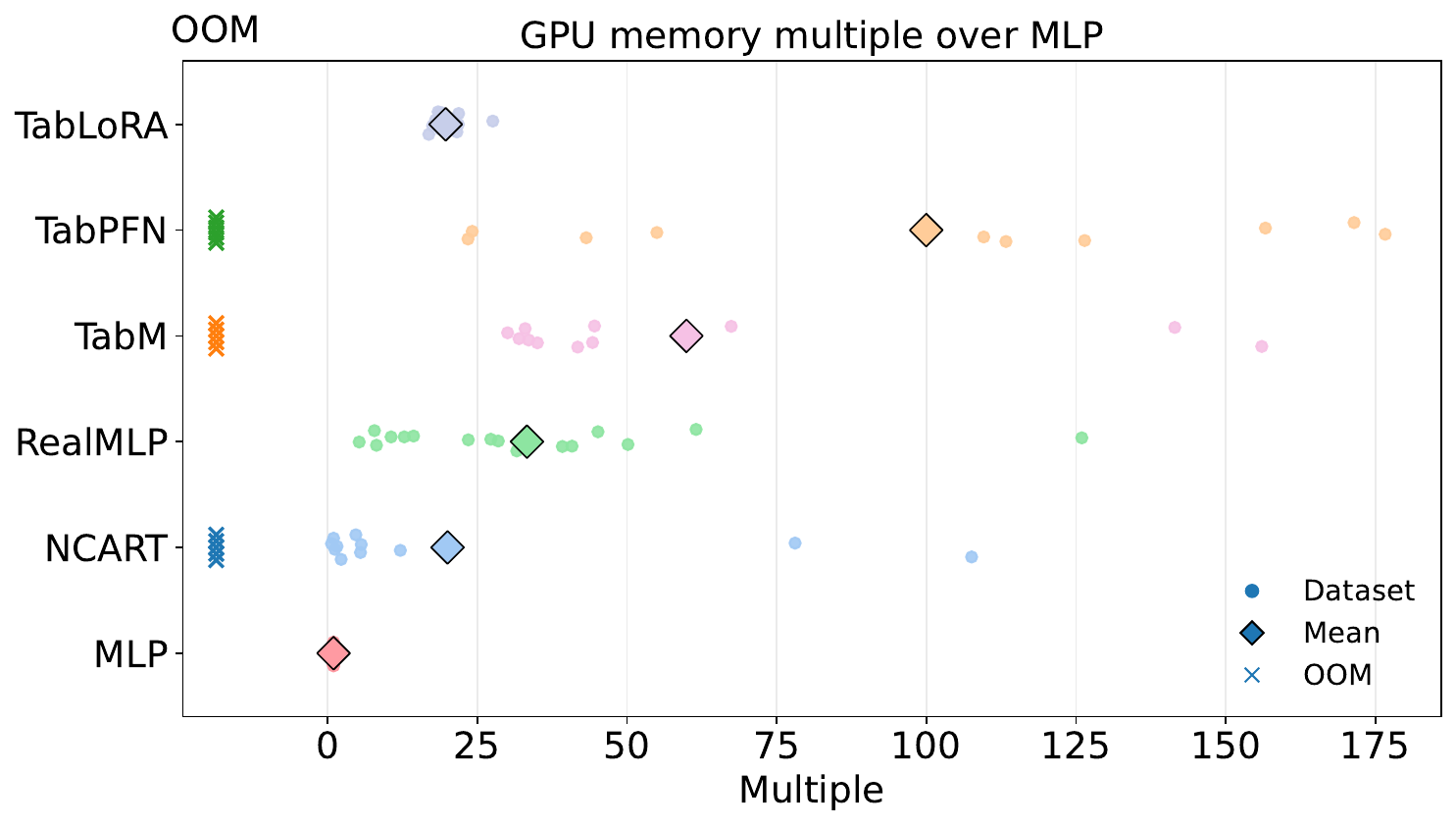}
    \caption{
    Dataset-level GPU memory multiple relative to MLP. 
    Each point represents one successful dataset-level run, and the diamond marker denotes the mean multiple over successful runs. 
    Cross markers indicate OOM cases, which are shown separately and are not assigned numerical memory multiples.
    }
    \label{f.gpu}
\end{figure}

Fig.~\ref{f.gpu} further shows the dataset-level GPU memory multiple relative to MLP. 
TabLoRA avoids OOM failures across all evaluated datasets and maintains a stable memory footprint. 
NCART uses relatively low memory on most successful runs, but fails on several datasets, suggesting that its memory behavior is less stable under large-scale settings. 
TabPFN and TabM show larger memory variation and multiple OOM cases, while RealMLP generally runs successfully but often requires more memory than TabLoRA. 
These results indicate that TabLoRA achieves a more reliable performance--memory trade-off under practical GPU memory constraints.

\begin{figure}[!ht]
    \centering
    \includegraphics[width=0.95\linewidth]{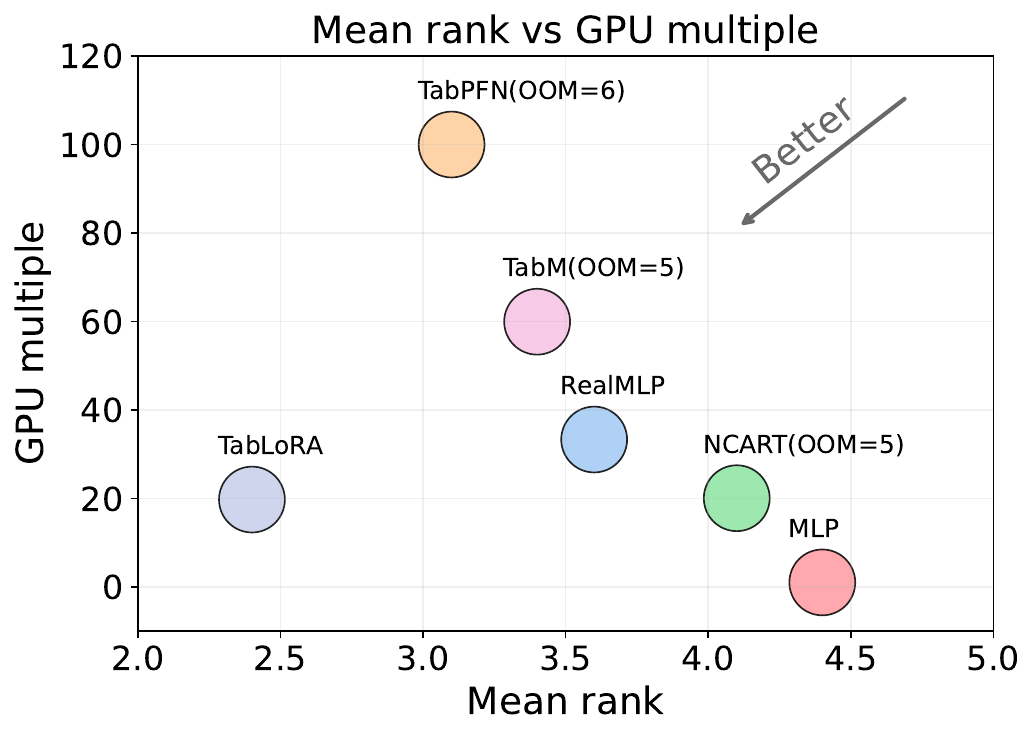}
    \caption{
    Performance--memory trade-off on large-scale datasets. 
    The x-axis reports the mean performance rank, and the y-axis reports the mean GPU memory multiple relative to MLP. 
    Lower values on both axes indicate better performance and lower GPU memory usage. 
    OOM counts are annotated next to the corresponding methods.
    }
    \label{f.bubble}
\end{figure}

\subsection{Ablation Study}

We conduct ablation studies to analyze the contributions of two key components in TabLoRA: the feature adapter and the low-rank ensemble parameterization. 
The first ablation evaluates whether the feature adapter improves the input representation. 
The second ablation examines whether the low-rank ensemble parameterization preserves the benefit of full ensembles while reducing parameter growth. 
The complete numerical results for all ablation experiments in this subsection are summarized in Table~\ref{T.ablation} in \ref{a.results}.

\subsubsection{Effect of Feature Adapter}

We first study the effect of the feature adapter. 
The variant without the feature adapter is referred to as \textit{Raw Input}, where all ensemble predictors receive the same original input representation. 
In contrast, the TabLoRA model uses the feature adapter to generate predictor-specific input representations before the shared backbone.

\begin{figure}[!ht]
    \centering
    \includegraphics[width=\linewidth]{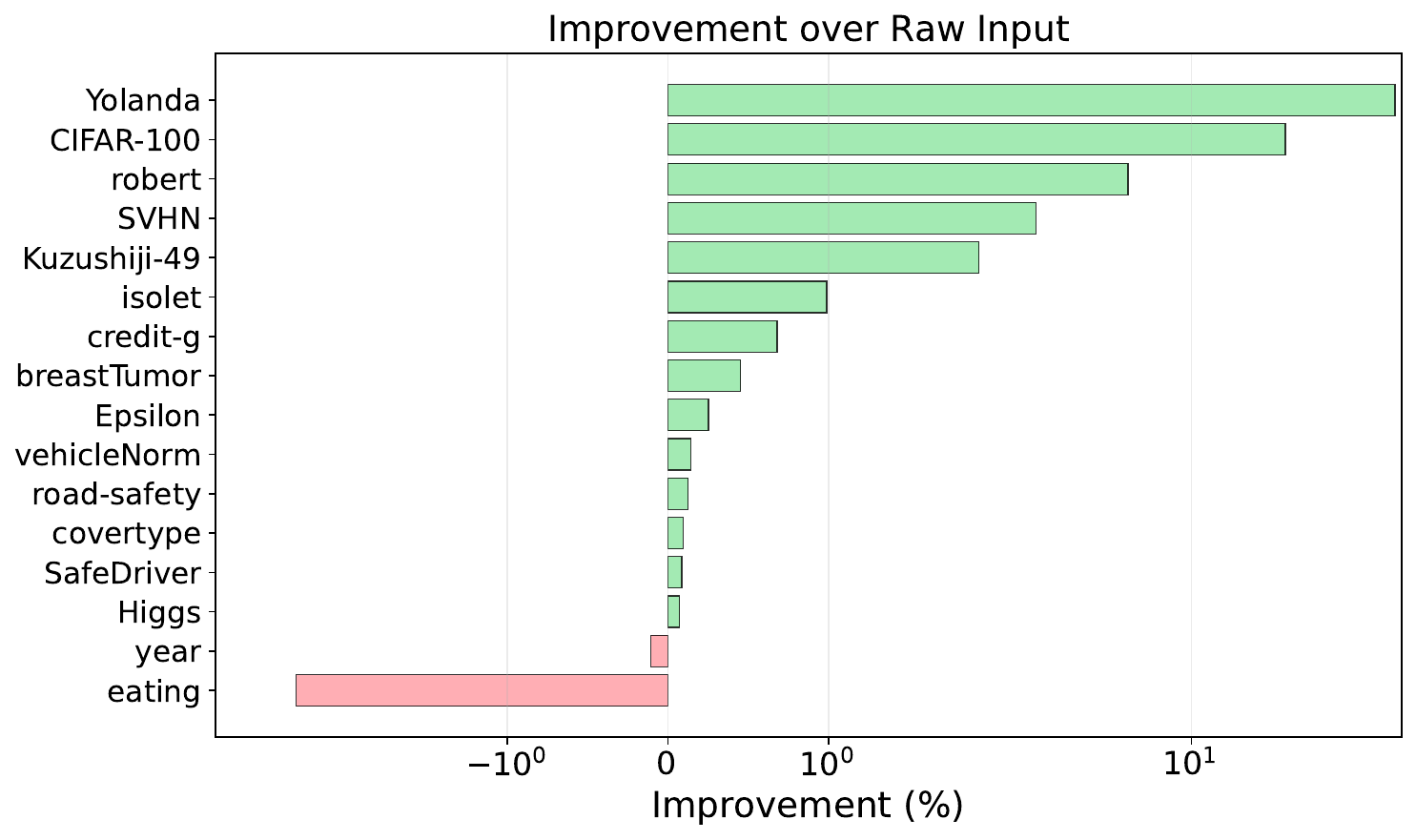}
    \caption{
    Dataset-level improvement of TabLoRA over Raw Input. 
    Positive values indicate that the feature adapter improves performance. 
    The x-axis uses a symmetric logarithmic scale to show both small and large improvements.
    }
    \label{f.adapter}
\end{figure}

Fig.~\ref{f.adapter} reports the dataset-level improvement of TabLoRA over Raw Input. 
Positive values indicate that the feature adapter improves performance, while negative values indicate degradation. 
The results show that the feature adapter improves performance on most datasets. 
Only two datasets show negative improvement, suggesting that the adapter is generally beneficial but still has dataset-dependent effects. 
Large improvements on datasets such as Yolanda, CIFAR-100, robert, and SVHN suggest that the feature adapter increases ensemble diversity by allowing different predictors to receive different input representations before the shared backbone.

\subsubsection{Effect of Low-Rank Ensemble Parameterization}

We next analyze the effect of the low-rank ensemble parameterization. 
We compare TabLoRA with two variants. 
\textit{Single} denotes a single predictor without ensemble modeling. 
\textit{Full} denotes a full deep ensemble, where each ensemble member has an independent backbone. 
TabLoRA lies between these two extremes: it shares the main backbone weights across predictors and introduces predictor-specific low-rank adaptations.

\begin{figure}[!ht]
    \centering
    \includegraphics[width=\linewidth]{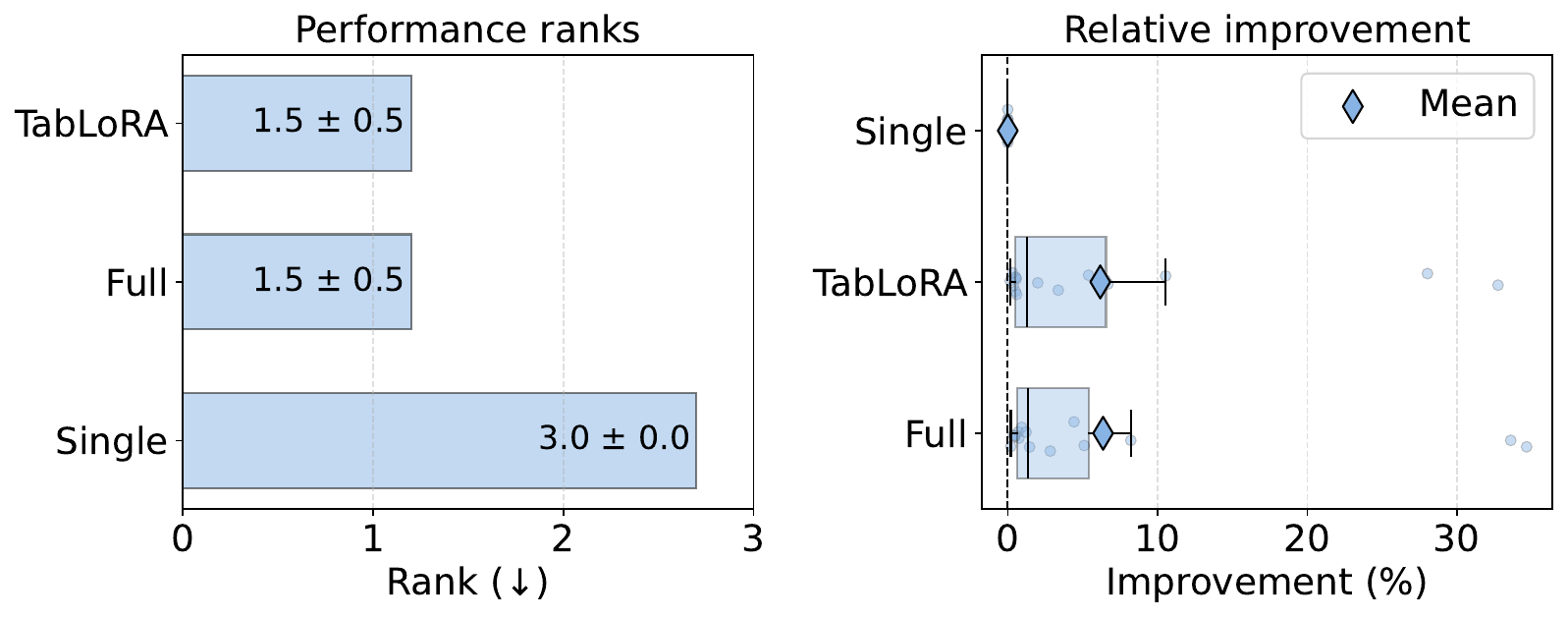}
    \caption{
    Comparison of Single, Full, and TabLoRA. 
    The left panel shows average ranks, and the right panel shows relative improvement over Single. 
    Diamonds denote mean improvements.
    }
    \label{f.lora}
\end{figure}

The first part of this ablation compares the predictive behavior of Single, Full, and TabLoRA after hyperparameter optimization (HPO). 
As shown in Fig.~\ref{f.lora}, Full and TabLoRA achieve the same average rank of $1.5 \pm 0.5$, while Single obtains the worst rank on every dataset. 
This indicates that ensemble-style modeling clearly improves over the single-predictor baseline. 
The relative improvement plot further shows that both Full and TabLoRA consistently improve over Single. 
Their improvement distributions are broadly comparable, indicating that TabLoRA can achieve full-ensemble-level predictive gains while avoiding full backbone duplication.

The second part isolates the structural parameter efficiency of TabLoRA. 
Since the HPO-selected configurations may use different hidden dimensions and depths, their parameter counts do not provide a clean architecture-level comparison. 
Therefore, we conduct a fixed-configuration parameter scaling analysis. 
We fix the input dimension to 500 and vary the hidden dimension and the number of backbone blocks. 
These correspond to the size of the hidden weights $W^\ell$ and the depth $L$ in Fig.~\ref{f.framework}. 
For each configuration, we compute the number of trainable parameters, measured in millions, for Full and TabLoRA, and report the parameter ratio of TabLoRA relative to Full.

\begin{figure}[!ht]
    \centering
    \includegraphics[width=\linewidth]{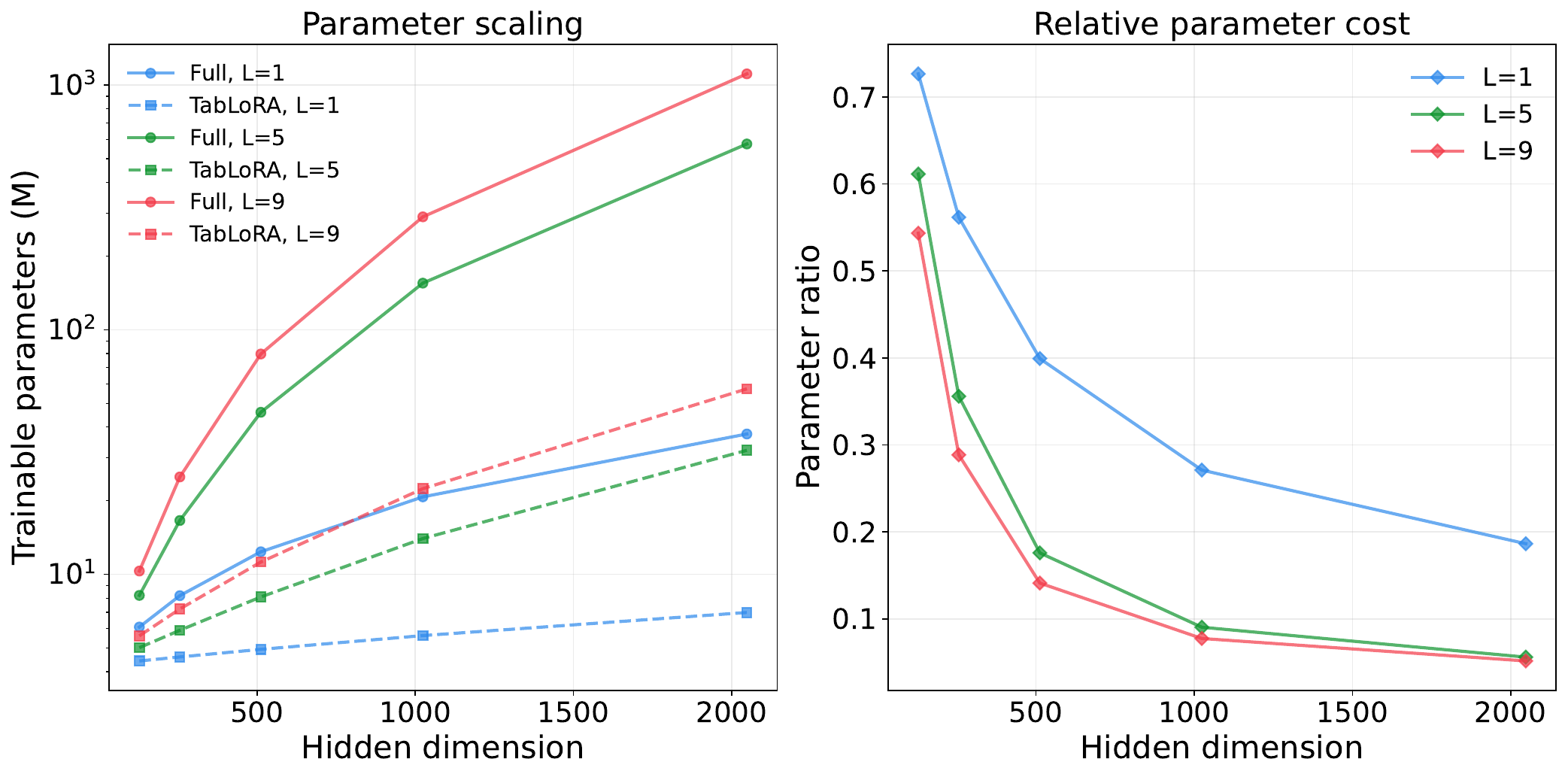}
    \caption{
    Fixed-configuration parameter scaling. 
    The left panel shows trainable parameters of Full and TabLoRA under different hidden dimensions and depths. 
    The right panel reports the parameter ratio of TabLoRA to Full.
    }
    \label{f.paramratio}
\end{figure}

Fig.~\ref{f.paramratio} shows that Full Ensemble parameters grow rapidly as the hidden dimension and depth increase. 
In contrast, TabLoRA grows much more slowly because it avoids duplicating the full backbone for each predictor. 
The right panel further confirms this trend: the parameter ratio of TabLoRA to Full decreases as the hidden dimension increases, especially for deeper networks. 
This supports the complexity analysis in \ref{s.complex} and shows that the low-rank ensemble design provides structural parameter efficiency.

\subsection{Performance on Small- and Medium-Scale Datasets}

\begin{figure}[!ht]
    \centering
    \includegraphics[width=\linewidth]{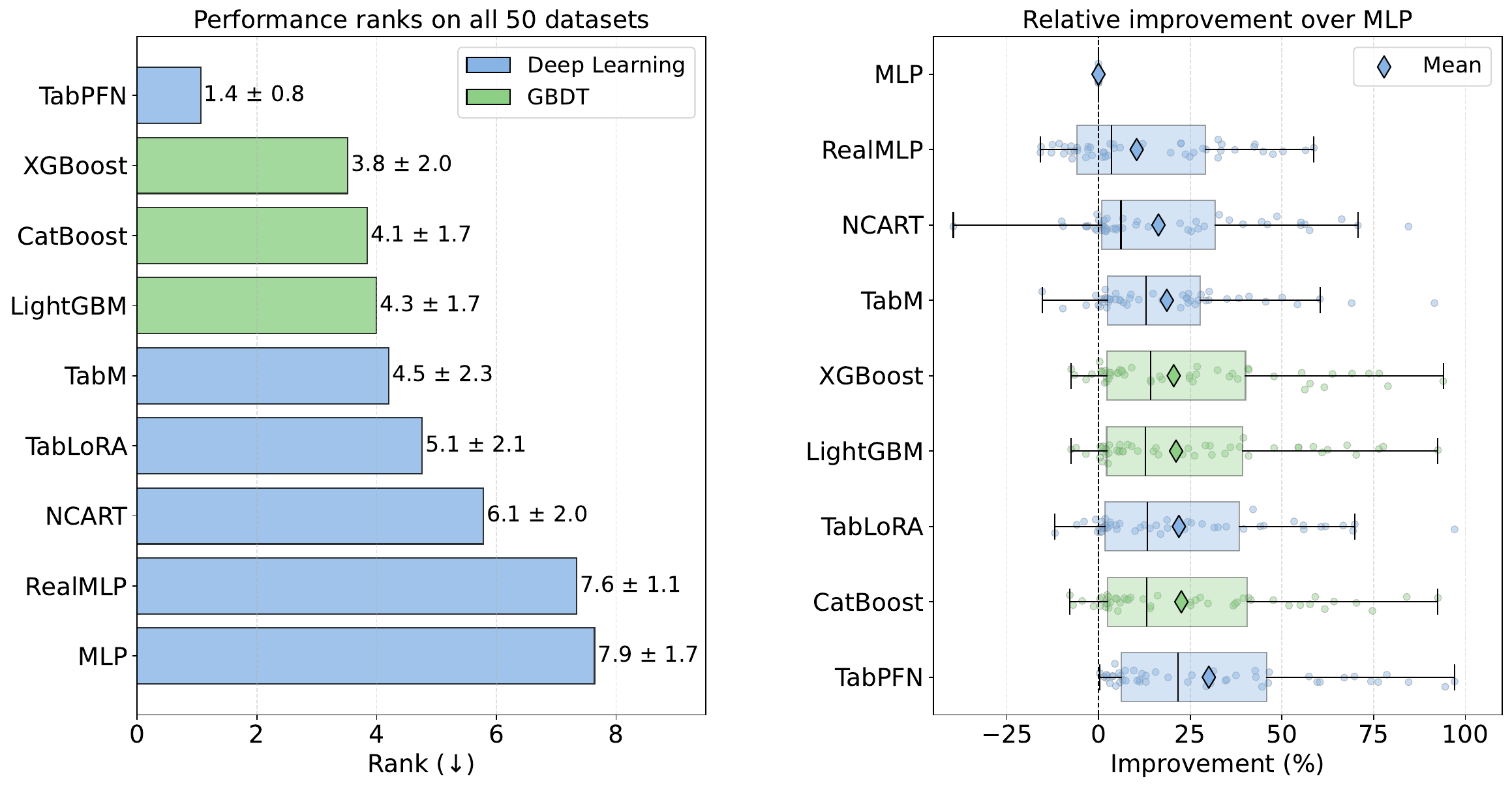}
    \caption{
    Performance on 50 small- to medium-scale datasets. 
    The left panel reports average ranks, where lower is better, and the right panel shows dataset-level relative improvement over MLP.
    }
    \label{f.small}
\end{figure}

To further evaluate the behavior of TabLoRA beyond large-scale settings, we conduct additional experiments on 50 small- to medium-scale datasets from CC18 \footnote{\url{https://www.openml.org/search?type=benchmark&sort=tasks_included&study_type=task&id=99}} and CTR23 \footnote{\url{https://www.openml.org/search?type=benchmark&sort=tasks_included&study_type=task&id=353}}. 
These datasets satisfy the following constraints: the number of samples is no more than 50,000, the number of classes is no more than 10 for classification tasks, and the number of features is no more than 2,000. 
Fig.~\ref{f.small} summarizes the average performance ranks and the relative improvement over MLP, while the complete numerical results are provided in Table~\ref{T.smallresults}.

The results show that TabPFN performs particularly well in this setting, achieving the best average rank among all compared methods. 
This is consistent with the design of TabPFN as a foundation-model approach for tabular prediction, especially in small-data regimes \cite{hollmann2022tabpfn,hollmann2025accurate}. 
The three GBDT methods also remain strong and stable baselines, with XGBoost, CatBoost, and LightGBM ranking immediately after TabPFN.

Compared with these strong baselines, TabLoRA does not achieve the best average rank. 
However, the relative improvement plot shows that TabLoRA still provides clear gains over the plain MLP baseline on many datasets. 
This indicates that the proposed low-rank ensemble design is beneficial beyond large-scale datasets, although its strongest advantage is not in outperforming TabPFN on small and medium-scale data. 
Instead, these results suggest that TabLoRA remains a competitive neural tabular model, while the main benefit of the method lies in scalable trainable ensemble learning and practical efficiency on large-scale datasets.

\section{Conclusions}
\label{s:conclu}

In this paper, we proposed TabLoRA, a parameter-efficient trainable neural ensemble framework for large-scale tabular learning. 
Motivated by the computational challenges of applying strong tabular models to large-scale datasets, TabLoRA constructs an ensemble using a shared backbone and predictor-specific low-rank adaptations. 
This design enables ensemble-style prediction without duplicating the full backbone for each predictor.
Extensive experiments show that TabLoRA maintains a stable GPU memory footprint while remaining competitive with strong GBDT methods, recent neural tabular models, and a pretrained tabular reference model. 
Ablation studies confirm the effectiveness of both the feature adapter and the low-rank ensemble parameterization, showing that TabLoRA can preserve much of the benefit of full ensembles while reducing parameter growth.
Overall, these results suggest that parameter-efficient ensemble design is a promising direction for practical tabular deep learning.
% Future work may further explore more adaptive low-rank structures, stronger feature adapters, and extensions of TabLoRA to broader tabular learning scenarios, such as missing-value handling, transfer learning, and online or streaming tabular prediction.

\section*{Declaration of generative AI use}
During the preparation of this work the authors used ChatGPT in order to improve the language and readability. After using this tool/service, the authors reviewed and edited the content as needed and take full responsibility for the content of the publication.

% \section*{Acknowledgment}
% The work is supported by the Key University Science Research Project of Jiangsu Province (No. 25KJB110014) and the Basic Research Program of Jiangsu (No. BK20250803).

\appendix
\section{Datasets Description}
\label{a.data}
Table. \ref{T.description} lists the datasets used in this paper, the column \#Target means the number of distinct values in the label. 
% The selection of data follows the following principles: 1. Diversity of feature types; 2. Diversity of feature dimensions; 3. Diversity of sample numbers.

\begin{table*}[!ht]
\renewcommand\arraystretch{1.1}
% \footnotesize
% \setlength\tabcolsep{10pt}
    \centering
    \caption{Details of datasets. \#Target in regression task means the number of unique values.}
    \label{T.description}
    \begin{adjustbox}{width=0.9\textwidth}
    \begin{tabular}{lcccc}
    \toprule[2pt]
        Dataset & \#Samples & \#Num.Feat. &\#Cat.Feat. & \#Target  \\
    \midrule[1.5pt]
        \multicolumn{5}{c}{Classification}\\
    \hline
        credit-g & 1000000  & 20  &13 & 2 \\
    \hline
        road-safety & 111762  & 32 &3 & 2  \\
    \hline
        Epsilon & 500000  & 2000 &0 & 2  \\
    \hline
        vehicleNorm & 98528  & 100 &0 & 2\\
    % \hline
    %     Diabetes130US & 71090  & 7 &0 & 2 \\
    \hline
        Higgs & 940160  & 24 &0 & 2\\
    % \midrule[1.5pt]
    %     \multicolumn{6}{c}{Multiclass Classification}\\
    \hline
        covertype & 581012  & 54 &44 & 7  \\
    \hline
        robert & 10000  & 7200 &0 & 10  \\
    \hline
        CIFAE-100 & 60000  & 3072 &0 & 100  \\
    \hline
        Kuzushiji-49 & 270912  & 784 &0 & 49   \\
    \hline
        isolet  & 7797  & 617 &0 & 26  \\
    \hline
        SVHN & 99289  & 3072 &0 & 10  \\
    % \hline
    %     Devnagari-Script & 92000  & 784 &0 & 46   \\
    \hline
        eating  & 945  & 6373 &0 & 7  \\
    \midrule[1.5pt]
    \multicolumn{5}{c}{Regression}\\
    \hline
        breastTumor & 116640  & 9 &8 & 1  \\
    \hline
        Yolanda & 400000  & 100 &0 & 1 \\
    \hline
        SafeDriver & 595212  & 223 &0 & 1  \\
    \hline
        year & 515345  & 90 &0 & 1  \\
    \bottomrule[2pt]
    \end{tabular}
    \end{adjustbox}
\end{table*}

\section{Optimization of hyperparameters}
\label{a.params}

Table.~\ref{T.hyperparam} lists the search range of hyperparameters, which refers to the original paper. 
We implement MLP and TabLoRA models using PyTorch and employ the official open-source implementations for other models 
\footnote{XGBoost: \url{https://xgboost.readthedocs.io/en/stable/}}
\footnote{CatBoost: \url{https://catboost.ai/}}
\footnote{LightGBM: \url{https://lightgbm.readthedocs.io/en/latest/}}
\footnote{RealMLP: \url{https://github.com/dholzmueller/pytabkit}}
\footnote{TabM: \url{https://github.com/yandex-research/tabm}}
\footnote{TabPFN: \url{https://github.com/PriorLabs/TabPFN}}.

\begin{table*}[!ht]
\footnotesize
\centering
\caption{Hyperparameters space.}
\label{T.hyperparam}
\begin{adjustbox}{width=\textwidth}
\begin{tabular}{ll|ll}
\toprule[2pt]
HyperParameters & Range & HyperParameters & Range \\
\midrule[1.5pt]
\multicolumn{4}{c}{XGBoost} \\
\hline
$num\_boost\_round$ & 200  & $early\_stopping\_rounds$ & 20  \\
\hline
$max\_depth$ & LogUniformInt [2, 10] & $alpha$ & LogUniform [1e-8, 0.1]  \\
\hline
$lambda$ & LogUniform [0.5, 2] & $eta$ & LogUniform [0.05, 0.3]  \\
\midrule[1.5pt]

\multicolumn{4}{c}{CatBoost} \\
\hline
$iterations$ & 200 & $od\_wait$ & 20\\
\hline
$max\_depth$ & LogUniformInt [2, 10] & $l2\_leaf\_reg$ & LogUniform [0.1, 2]\\
\hline
$learning\_rate$ & LogUniform [0.05, 0.3] &&\\
\midrule[1.5pt]

\multicolumn{4}{c}{LightGBM} \\
\hline
$iterations$ & 200 & $early\_stopping\_round$ & 20\\
\hline
$num\_leaves$ & LogUniformInt [8, 48] & $lambda\_l_1$ & LogUniform [1e-8, 0.1]\\
\hline
$lambda\_l_2$ & LogUniform [1e-8, 0.1] & $learning\_rate$ & LogUniform [0.05, 0.3] \\
\midrule[1.5pt]

\multicolumn{4}{c}{MLP} \\
\hline
$n\_layers$& UniformInt [1, 9] & $d\_hidden$& UniformInt [64, 512], step=32  \\
\hline
$dropout$& [0, 0.5] && \\
\midrule[1.5pt]

\multicolumn{4}{c}{RealMLP} \\
\hline
$n\_blocks$& UniformInt [1, 10] & $hidden\_sizes$& [64, 256, 512]  \\
\hline
$dropout$& [0, 0.15, 0.3] && \\
\midrule[1.5pt]

\multicolumn{4}{c}{NCART} \\
\hline
$n\_trees$  & [8, 16, 32, 64] & $n\_selected$ & UniformInt [2, 10]  \\
\hline
$n\_layers$ & [2, 4]  & $mask\_type$ & \ [sparsemax, entmax]  \\
\midrule[1.5pt]

\multicolumn{4}{c}{TabM} \\
\hline
$n\_bins$  & UniformInt [2, 128] & $d\_embedding$ & UniformInt [8, 32],step=4  \\
\hline
$n\_blocks$ & UniformInt [1, 4]  & $d\_blocks$ & \ UniformInt [64, 1024],step=16  \\
\hline 
$dropout$& [0, 0.5] & $k$& 32\\
\midrule[1.5pt]

\multicolumn{4}{c}{TabLoRA} \\
\hline
$d\_hidden$  & UniformInt [64, 512], step=32 & $dropout$& [0, 0.5] \\
\hline
$n\_blocks$ & UniformInt [1, 9]  & $r\_rank$ & \ UniformInt [4, 16],step=2  \\
\hline 
$K$& 32 & &\\

\bottomrule[2pt]
\end{tabular}
\end{adjustbox}
\end{table*}

\section{More results}
\label{a.results}

\begin{table*}[!ht]
\renewcommand\arraystretch{1.2}
    \centering
    \caption{Practical memory usage (MB). The \textbf{bold} indicates the top result; \textit{OOM} represents there exists GPU overflow}
    \label{T.memoryresults}
    \begin{adjustbox}{width=\textwidth}
    \begin{tabular}{lcccccc}
    \toprule[2pt]
        Dataset  & MLP & RealMLP & NCART & TabM & TabPFN & TabLoRA \\
    \midrule[1.5pt]
        credit-g & \textbf{103.2} & 6351.6 & 103.2 & 3466.4 & 18229.6 & 1948.0 \\
        road-safety & \textbf{108.4} & 1150.0 & 171.2 & 3466.0 & 2541.6 & 1954.4 \\
        Epsilon & \textbf{128.4} & 16174.8 & \textit{OOM} & \textit{OOM} & \textit{OOM} & 2773.6 \\
        vehicleNorm & \textbf{106.0} & 1357.2 & 582.4 & 4726.0 & 2560.8 & 1973.6 \\
        Higgs & \textbf{102.4} & 5137.2 & 126.8 & 3588.8 & 17556.0 & 1957.6 \\
        covertype & \textbf{117.2} & 3703.2 & 265.2 & 3523.6 & 12844.0 & 1981.6 \\
        robert & \textbf{215.2} & 3090.8 & \textit{OOM} & \textit{OOM} & \textit{OOM} & 5938.4 \\
        CIFAR-100 & \textbf{151.6} & 4322.8 & \textit{OOM} & \textit{OOM} & \textit{OOM} & 3318.0 \\
        Kuzushiji-49 & \textbf{128.0} & 5018.8 & 13768.4 & 19971.6 & 14503.2 & 2244.4 \\
        isolet & \textbf{118.8} & 626.8 & 9276.4 & 16810.0 & 6533.6 & 2234.4 \\
        SVHN & \textbf{148.8} & 6720.0 & \textit{OOM} & \textit{OOM} & \textit{OOM} & 3248.8 \\
        eating & \textbf{218.4} & 1708.4 & \textit{OOM} & \textit{OOM} & \textit{OOM} & 4472.8 \\
        breastTumor & 105.2 & 859.2 & \textbf{69.2} & 3471.6 & 4544.4 & 1961.2 \\
        Yolanda & \textbf{106.8} & 2510.0 & 600.4 & 4726.8 & 13504.4 & 1971.6 \\
        SafeDriver & \textbf{105.2} & 4294.8 & 1278.4 & 7092.4 & \textit{OOM} & 2014.8 \\
        year & \textbf{108.4} & 2955.2 & 511.6 & 4528.4 & 16979.6 & 1967.6 \\
    \midrule[1.5pt]
    Mean &  \textbf{129.5}  & 4123.8  & 2432.11 & 6851.96 & 10979.72  & 2622.55    \\
    Best/Worst & \textbf{15/0} & 0/0 & 1/5 & 0/5 & 3/9 & 0/0\\
    \bottomrule[2pt]
    \end{tabular}
    \end{adjustbox}
\end{table*}

% \begin{table*}[!ht]
% \renewcommand\arraystretch{1.1}
% % \setlength\tabcolsep{10pt}
%     \centering
%     \caption{Mean $\pm$ std. results of 9 models on 16 datasets. The \textbf{bold} indicates the top result.}
%     \label{T.mainimprov}
%     \begin{adjustbox}{width=\textwidth}
%     \begin{tabular}{lccccccccc}
%     \toprule[2pt]
%         Dataset & XGBoost & CatBoost & LightGBM & MLP & RealMLP & NCART & TabM & TabPFN & TabLoRA \\
%     \midrule[1.5pt]
%         \multicolumn{10}{c}{Binary Classification (AUC $\uparrow$)}\\

%     \midrule[1.5pt]
%         \multicolumn{10}{c}{Multiclass Classification (Acc. $\uparrow$)}\\

%     \midrule[1.5pt]
%     \multicolumn{10}{c}{Regression (MSE $\downarrow$)}\\

% \midrule[1.5pt]
% Mean rank & 3.9 & 4.2 & 4.3& 7.9& 7.6& 6.1& 4.5& \textbf{1.4}& 5.1 \\
% Best/Worst & 3/2 & 1/0 & 1/0 & 0/26 & 0/13 & 2/5 & 2/2 & \textbf{37/0} & 2/2\\
%     \bottomrule[2pt]
%     \end{tabular}
%     \end{adjustbox}
% \end{table*}

\begin{table*}[!ht]
\renewcommand\arraystretch{1.1}
    \centering
    \caption{Mean results of the ablation studies. 
    \textit{Raw} denotes the TabLoRA variant without the feature adapter, and \textit{Improv.} denotes the improvement of TabLoRA over Raw. \textit{Single} denotes a single-predictor model, and \textit{Full} denotes the full deep ensemble with independent predictors. }
    \label{T.ablation}
    \begin{adjustbox}{width=0.9\textwidth}
    \begin{tabular}{lc||cc||cc}
    \toprule[2pt]
    
        Dataset & TabLoRA & Raw & Improv. & Single & Full \\
    \midrule[1.5pt]
        \multicolumn{6}{c}{Binary Classification (AUC $\uparrow$)}\\
        credit-g & 88.39 & 87.80 & 0.68 & 87.96 & 88.57 \\
        road-safety & 88.34 & 88.23 & 0.13 & 87.85 & 88.50 \\
        Epsilon & 96.23 & 95.99 & 0.25 & 95.88 & 96.24 \\
        vehicleNorm & 92.43 & 92.30 & 0.14 & 91.97 & 92.40 \\
        Higgs & 83.76 & 83.70 & 0.07 & 83.62 & 84.39 \\
    \midrule[1.5pt]
        \multicolumn{6}{c}{Multiclass Classification (Acc. $\uparrow$)}\\
        covertype & 96.05 & 95.96 & 0.09 & 95.49 & 96.65 \\
        robert & 44.18 & 41.72 & 6.04 & 40.06 & 43.27 \\
        CIFAR-100 & 30.70 & 25.52 & 21.13 & 23.38 & 31.13 \\
        Kuzushiji-49 & 93.63 & 91.87 & 1.94 & 90.61 & 94.58 \\
        isolet & 97.06 & 96.12 & 0.99 & 95.18 & 96.55 \\
        SVHN & 88.78 & 86.31 & 2.90 & 84.28 & 88.53 \\
        eating & 59.47 & 61.27 & -2.99 & 46.67 & 62.01 \\
    \midrule[1.5pt]
    \multicolumn{6}{c}{Regression (MSE $\downarrow$)}\\
        breastTumor & 87.68 & 88.08 & 0.45 & 88.16 & 87.74 \\
        Yolanda & 79.20 & 160.20 & 50.56 & 84.88 & 79.43 \\
        SafeDriver & 0.03 & 0.03 & 0.09 & 0.03 & 0.03 \\
        year & 75.49 & 75.42 & -0.10 & 80.76 & 78.47 \\
    \bottomrule[2pt]
    \end{tabular}
    \end{adjustbox}
\end{table*}

\begin{table*}[!ht]
\renewcommand\arraystretch{1.1}
    \centering
    \caption{Mean results of 9 models on 50 datasets. The \textbf{bold} indicates the top result.}
    \label{T.smallresults}
    \begin{adjustbox}{width=\textwidth}
    \begin{tabular}{lccccccccc}
    \toprule[2pt]
        Dataset & XGBoost & CatBoost & LightGBM & MLP & RealMLP & NCART & TabM & TabPFN & TabLoRA \\
    \midrule[1.5pt]
        \multicolumn{10}{c}{Binary Classification (AUC $\uparrow$)}\\
ilpd & 74.82 & 74.40 & 73.59 & 72.71 & 67.24 & 73.33 & 73.37 & \textbf{77.08} & 72.53 \\
bank-marketing & 93.46 & 93.36 & 93.75 & 88.74 & 92.31 & 90.78 & 93.99 & \textbf{94.57} & 93.13 \\
churn & 91.36 & 91.87 & 91.43 & 65.18 & 84.03 & 90.45 & 91.51 & \textbf{92.60} & 90.56 \\
credit-g & 76.01 & 77.61 & 76.29 & 55.37 & 69.46 & 74.76 & 68.52 & \textbf{79.59} & 78.27 \\
dresses-sales & 60.55 & 58.90 & 60.05 & 51.76 & 53.37 & 51.30 & 65.55 & 67.67 & \textbf{67.77} \\
cylinder-bands & 87.82 & 83.06 & 86.09 & 50.98 & 76.11 & 84.11 & 63.25 & \textbf{88.09} & 81.91 \\
adult & 92.87 & 92.76 & \textbf{92.89} & 63.14 & 91.05 & 90.78 & 91.47 & 92.05 & 91.11 \\
tic-tac-toe & 99.92 & \textbf{99.99} & 99.68 & 94.50 & 82.51 & 99.80 & 99.96 & 99.94 & 99.90 \\
credit-approval & 93.59 & 93.74 & 93.20 & 68.82 & 91.62 & 91.14 & 84.41 & \textbf{94.23} & 92.44 \\
climate-model & \textbf{94.01} & 90.76 & 92.90 & 91.98 & 89.43 & 93.38 & 93.66 & 93.96 & 92.17 \\
diabetes & 82.03 & 81.61 & 81.29 & 70.42 & 74.52 & 82.20 & 83.28 & \textbf{83.59} & 81.36 \\
pc1 & 86.76 & 84.73 & 85.59 & 48.94 & 69.37 & 82.83 & 65.74 & \textbf{89.46} & 82.44 \\
qsar-biodeg & 92.35 & 91.75 & 92.46 & 91.32 & 88.22 & 92.15 & 93.00 & \textbf{93.48} & 91.78 \\
phoneme & 95.36 & 95.36 & 95.17 & 93.31 & 91.28 & 90.72 & 94.75 & \textbf{97.43} & 94.06 \\
pc3 & 82.31 & 82.29 & 81.92 & 76.91 & 71.96 & 81.90 & 69.54 & \textbf{85.93} & 84.48 \\
blood-transfusion & 73.14 & 74.83 & 72.05 & 58.81 & 75.27 & 74.20 & 76.20 & 76.71 & \textbf{77.66} \\
wdbc & 99.34 & 99.54 & 99.09 & 79.80 & 99.40 & 99.74 & 99.56 & \textbf{99.85} & 99.72 \\
MagicTelescop & 93.49 & 93.34 & 93.35 & 88.92 & 91.19 & 92.94 & 93.06 & \textbf{95.48} & 93.13 \\
jm1 & 73.79 & 72.52 & 73.34 & 69.36 & 70.38 & 72.04 & 70.91 & \textbf{77.05} & 71.57 \\
kc2 & 82.92 & 80.65 & 82.03 & 53.43 & 70.54 & \textbf{83.60} & 81.87 & 83.51 & 82.77 \\
kc1 & 80.23 & 79.69 & 79.80 & 77.78 & 75.56 & 79.28 & 77.76 & \textbf{85.17} & 79.44 \\
ozone-level & 91.77 & 91.46 & 91.23 & 72.52 & 86.59 & 91.96 & \textbf{93.43} & 94.37 & 92.73 \\
pc4 & 93.51 & 93.29 & 92.97 & 60.54 & 88.84 & 93.39 & 76.26 & \textbf{96.02} & 92.28 \\
    \midrule[1.5pt]
        \multicolumn{10}{c}{Multiclass Classification (Acc. $\uparrow$)}\\
dna & 96.18 & 96.14 & 96.30 & 93.57 & 85.05 & 95.70 & 96.11 & \textbf{97.15} & 94.86 \\
eucalyptus & 64.60 & 67.30 & 65.00 & 37.03 & 48.78 & 57.30 & 61.89 & \textbf{71.08} & 62.03 \\
cmc & \textbf{56.81} & 55.80 & 55.05 & 49.83 & 44.61 & 54.98 & 55.19 & 56.14 & 50.71 \\
splice & 95.80 & 95.80 & 96.43 & 84.01 & 93.35 & 89.50 & 96.36 & \textbf{96.93} & 95.64 \\
junglechess & 86.50 & 86.19 & 86.93 & 92.52 & 83.87 & 83.69 & \textbf{99.66} & 96.08 & 87.11 \\
steel-plates-fault & 80.26 & 80.82 & 79.64 & 47.87 & 65.30 & 73.73 & 40.67 & \textbf{83.60} & 76.30 \\
semeion & 90.91 & 91.98 & 90.60 & 90.22 & 83.82 & 91.54 & 92.92 & \textbf{95.24} & 93.10 \\
segment & 92.86 & 92.38 & 92.47 & 89.96 & 84.85 & 90.00 & 92.86 & \textbf{94.16} & 92.21 \\
mfeat-zernike & 77.95 & 77.65 & 77.95 & 84.15 & 71.15 & 81.25 & 83.55 & \textbf{84.45} & 83.50 \\
mfeat-morphological & 73.55 & 73.00 & 73.00 & 38.40 & 60.35 & 70.00 & 72.65 & \textbf{74.70} & \textbf{74.70} \\
mfeat-fourier & 82.50 & 83.70 & 82.75 & 81.15 & 71.10 & 82.55 & 84.90 & \textbf{90.10} & 82.70 \\
GesturePhase & 66.91 & 66.22 & 66.23 & 47.76 & 53.41 & 52.55 & 65.66 & \textbf{80.38} & 56.56 \\
vehicle & 76.00 & 76.94 & 76.24 & 71.53 & 60.35 & 74.94 & 76.35 & \textbf{88.71} & 80.35 \\
wall-robot-navigation & \textbf{99.71} & 99.45 & 99.69 & 91.52 & 92.64 & 93.41 & 99.25 & 98.10 & 92.40 \\
satimage & 91.80 & 91.56 & 91.62 & 91.15 & 89.58 & 90.70 & 91.54 & \textbf{93.94} & 87.62 \\
    \midrule[1.5pt]
    \multicolumn{10}{c}{Regression (MSE $\downarrow$)}\\
cps88wages & 139678 & 138929 & 138974 & 140145 & 150520 & 140697 & 138459 & \textbf{138024} & 138910 \\
healthinsurance & 209.84 & 209.17 & 209.40 & 213.82 & 211.43 & 211.97 & \textbf{208.54} & 209.22 & 209.84 \\
studentperformancepor & 8.33 & 8.40 & 7.81 & 11.22 & 10.74 & 15.65 & 7.83 & \textbf{7.34} & 8.49 \\
forestfires & 1992 & 1795 & 1801 & 1957 & 1920 & \textbf{1690} & 1783 & 1704 & 1733 \\
abalone & 4.48 & 4.53 & 4.48 & 4.34 & 4.59 & 4.47 & 4.48 & \textbf{3.88} & 4.85 \\
kin8nm & 0.0151 & 0.0093 & 0.0128 & 0.0055 & 0.0109 & 0.0088 & 0.0046 & \textbf{0.0039} & 0.0054 \\
pumadyn32nh & 0.0005 & 0.0005 & 0.0005 & 0.0011 & 0.0007 & 0.0008 & \textbf{0.0004} & \textbf{0.0004} & 0.0005 \\
whitewine & 0.4085 & 0.3955 & 0.4160 & 0.6352 & 0.4846 & 0.4939 & 0.4929 & \textbf{0.3412} & 0.5164 \\
physiochemicalprotein & 13.62 & 14.41 & 14.11 & 37.59 & 16.37 & 19.30 & 27.08 & \textbf{8.05} & 12.52 \\
superconductivity & 93.04 & 98.99 & 94.52 & 242.05 & 139.33 & 130.54 & 120.85 & \textbf{79.94} & 135.34 \\
QSARfishtoxicity & 0.9785 & 0.9364 & 0.9456 & 1.3345 & 1.0349 & 1.4682 & 1.0139 & \textbf{0.8679} & 1.1087 \\
redwine & 0.3494 & 0.3247 & 0.3574 & 0.5171 & 0.4008 & 0.3909 & 0.3904 & \textbf{0.2952} & 0.3957 \\
\midrule[1.5pt]
Mean rank & 3.8 & 4.1 & 4.3& 7.9& 7.6& 6.1& 4.5& \textbf{1.4}& 5.1 \\
Best/Worst & 3/2 & 1/0 & 1/0 & 0/26 & 0/13 & 2/5 & 3/2 & \textbf{39/0} & 2/2\\
    \bottomrule[2pt]
    \end{tabular}
    \end{adjustbox}
\end{table*}

\bibliographystyle{elsarticle-num} 
\bibliography{references}

\end{document}